\documentclass{article}
\usepackage{spconf,amsmath,graphicx}
\usepackage{balance}
\usepackage{bibspacing}
\usepackage{color}
\usepackage{enumitem}
\usepackage[font={footnotesize}]{caption}
\usepackage{multirow}
\usepackage{subfigure}
\usepackage{url}

\setlist{nolistsep}

\title{\textsc{The SWAX Benchmark:}\\Attacking Biometric Systems with Wax Figures}
\name{
  Rafael Henrique Vareto, Araceli Marcia Sandanha, William Robson Schwartz
  }
\address{
  Smart Sense Laboratory, Department of Computer Science\\
  Universidade Federal de Minas Gerais, Belo Horizonte, Brazil\\
}
\newcommand{\datasetAbbr}{\textsc{swax}}
\newcommand{\datasetName}{Sense Wax Attack dataset}
\newcommand{\etal}{~\textit{et~al.}~}

\begin{document}
  \ninept
  
  \maketitle
  \begin{abstract}
	A face spoofing attack occurs when an intruder attempts to impersonate someone who carries a gainful authentication clearance.
	It is a trending topic due to the increasing demand for biometric authentication on mobile devices, high-security areas, among others.
	This work introduces a new database named \datasetName~(\datasetAbbr), comprised of real human and wax figure images and videos that endorse the problem of face spoofing detection.
	The dataset consists of more than 1800 face images and 110 videos of 55 people/waxworks, arranged in training, validation and test sets with a large range in expression, illumination and pose variations.
	Experiments performed with baseline methods show that despite the progress in recent years, advanced spoofing methods are still vulnerable to high-quality violation attempts.
\end{abstract}

  \section{Introduction}
\label{sec:intro}


The term \emph{spoofing}, also referred to as \emph{presentation attack}, represents a legitimate threat for biometric systems.
The attack occurs when a malefactor attempts to pass him/herself off as someone who carries an advantageous authentication clearance.
Intruders regularly employ falsified data to bypass security procedures and gain illegitimate access. 
As a countermeasure, new datasets are regularly released in the literature as an attempt to address the latest attack nuances and leverage upcoming researches~\cite{chingovska2012effectiveness,pinto2015face,bhattacharjee2018spoofing}.

The human face may have become the ``universal'' authentication biometric for several reasons, such as the increasing distribution of personal pictures on social networks, spread of surveillance cameras, and convenience to name a few~\cite{kumar2017comparative}.
Most spoofing attacks comprise facial pictures from users enrolled in recognition systems since face images are easily acquired due to their broad availability.
{A case in point occurred in China involving scammers posing with wax figures of prominent executives to lure in around 600K investors and embezzle almost US\$500 million~\cite{china2012scam}}.
The low-cost access to face images contributes to the increase of criminals designing attacks to be validated as authentic users, turning face spoofing into a popular way of deceiving people and biometric applications.

\begin{figure}
    \centering
    \includegraphics[width=0.73\columnwidth]{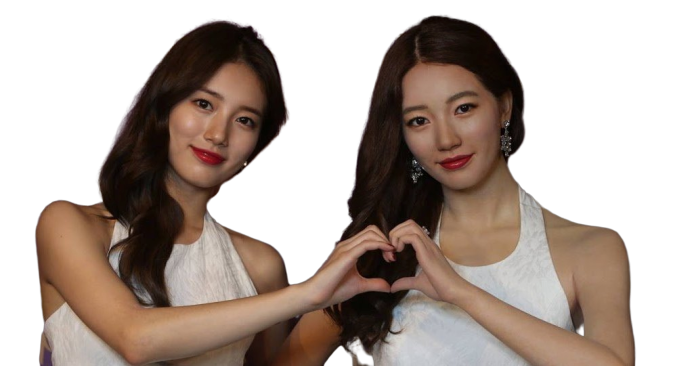}
    \caption{A Korean celebrity on the left and a realistic sculpted wax dummy.}
    \vspace{-4mm}
    \label{fig:fake-real}
\end{figure}

This work presents a database comprised of both real human and wax figure image/video in the interest of endorsing the problem of face spoofing detection.
The proposed database, entitled \datasetName~(\datasetAbbr), is designed to investigate the problem in which a face media is presented to a system that must determine whether it categorizes a \textit{bona fide} (authentic) or a counterfeit (attack) sample.
{The goal is not only to deliver a novel spoofing benchmark by specifying dataset protocols and usage requirements, but also enable the development of more robust anti-spoofing systems with the anticipation of unforeseen wax-based portrait attacks.}

Although several datasets focus on spoofing detection, most engage in recapturing authentic images or videos in distinct mediums~\cite{wen2015face,boulkenafet2017oulu,liu2018learningSIW} by varying input sensors, attack types and capture conditions.
Only recently have few researchers turned themselves to modeling emerging attack strategies~\cite{liu20163d,agarwal2017face,manjani2017detecting}, e.g. silicon masks, which suggests concealing a suspect's real appearance or impersonating someone else's identity. 
To the best of our knowledge, \datasetAbbr~is the only dataset comprising both real and wax-modeled persons up to the present time.
It provides means of studying attacks in unconstrained environments as it encompasses significant divergence in pose, expression, lighting, scene, and camera settings.

The proposed \datasetAbbr~dataset consists of genuine and counterfeit samples for all available subjects.
As illustrated in Figure~\ref{fig:fake-real}, the database contains labeled photographs of characters, celebrities and public figures, chosen based on a list of waxworks obtained from a famous chain of wax museums.
More precisely, this work aims at investigating face spoofing detection in realistic scenarios, whither there is little control over the images acquisition.
In contemplation of fair algorithm comparisons, we provide four protocols for developing and evaluating algorithms using the \datasetAbbr~benchmark.

The main contributions of this work are: 
\emph{1)}~generation of a public set of real human and wax figure media \textit{in the wild}, indicating that these faces carry most common variations observed in ordinary scenarios and, consequently, represent practical situations; 
\emph{2)}~development of straightforward baselines, which stand on long-established techniques in the interest of meeting the requirements of each proposed protocol and guarantee a legitimate comparison among algorithms considering the sort of data they are trained on.


\section{Related works}
\label{sec:related}

\begin{figure*}[!t]
    \centering
    \includegraphics[width=0.90\linewidth]{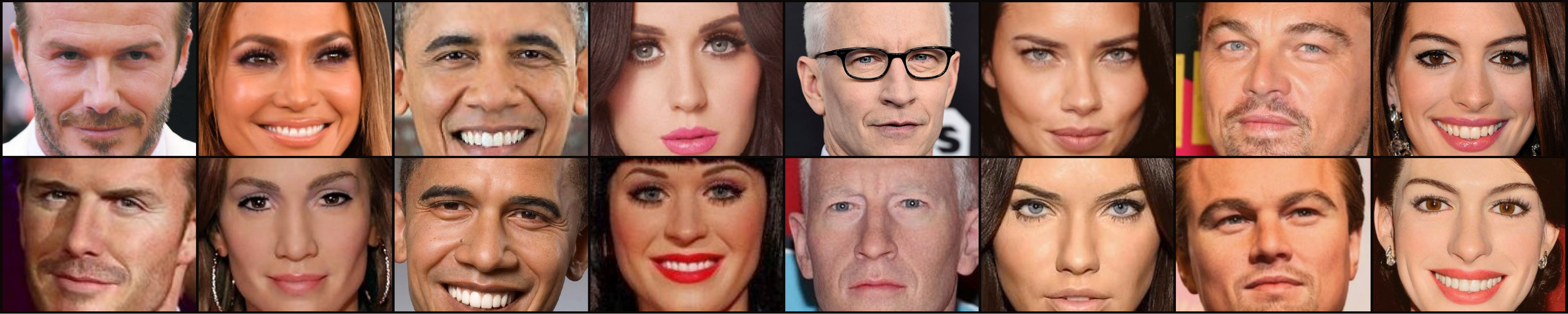}
    \caption{Face samples collected from online resources and constituting \textit{bona fide} (top row) and \textit{counterfeit} (bottom row) images.}
    \vspace{-4mm}
    \label{fig:bona-wax-faces}
\end{figure*}

Most existing datasets concentrate either on modeling impersonation attack strategies or recapturing genuine images and videos on different mediums.
The former demands a high expertise level since it requires the manufacturing of realistic masks and, therefore, is an expensive process.
The latter usually involves high-quality cameras for the sake of capturing deceptive print and video replay attacks.

Several face spoofing datasets have been proposed in the last decade~\cite{zhang2012face, chingovska2012effectiveness, wen2015face}.
More recently, Boulkenafet\etal\cite{boulkenafet2017oulu} designed one of the largest mobile-based benchmarks, \textsc{oulu-npu}, holding more than four thousand real-access and attack videos captured using the frontal camera of six mobile devices.
Liu\etal\cite{liu2018learningSIW} introduced \textsc{siw}, with live and spoof videos of 165 individuals covering a large range of expression, illumination and pose variations.
{These datasets represent situations that fail to generalize well in conditions where attacks do not proceed from print or replay spoof mediums. 
More precisely, they require that algorithms learn whether a medium corresponds to an attack rather than the individual itself.
A way to make biometric systems robust to other realistic intrusions is to lay out medium-free presentation attacks, like mask-based attacks.}

Few masks-based attacks have also been proposed~\cite{erdogmus2013spoofing, liu20163d}.
Manjani\etal\cite{manjani2017detecting} propose \textsc{smad}, a database containing stretchable masks under unconstrained environments, allowing actors to speak and blink, in an attempt to pass on a lifelike sensation.  
Bhattacharjee\etal\cite{bhattacharjee2018spoofing} consider rigid and flexible silicone masks as they release \textsc{cs-mad}, a dataset captured using low-cost cameras with the collaboration of 14 subjects.
{The use of masks can restrict natural facial functions/movements such as smiling, talking and blinking, failing to proper fit and match real faces.
It becomes even more complex under the need of accurately reproducing another person's appearance due to the requirement of facial casts.
Realistic mask impersonations demand face molds from the subject being modeled and from the person about to wear the disguise.
The process of making lifelike masks can cost high figures due to its manufacturing complexity and, to make matters worse, real presentation attacks do not rely on the cooperation of the individual ``being cloned''.} 


There are numerous approaches focusing on print or replay spoofing attacks. 
Many methods deal with the design of handcrafted descriptors and learning algorithms whereas others focus on the convolutional neural networks trend, described as follows.

There is a myriad of anti-spoofing works in the literature~\cite{wen2015face, pinto2015using, li2016original}.
Valle\etal\cite{valle2017transfer} presented a transfer learning method using a pre-trained \textsc{dnn} model on static features to recognize photo, video and mask attacks.
Liu\etal\cite{liu2018learningSIW} combined \textsc{dnn} with \textsc{rnn} to estimate the depth of face images along with r\textsc{ppg} signals to boost the detection of unauthorized access.
Jourabloo\etal\cite{jourabloo2018face} decomposed a spoofing face into noise and face information by modeling the process of how the attack is generated from its original live image. 

Only recently have researchers turned their attention to fraudulent 3D-mask attacks. 
Steiner\etal\cite{steiner2016design} adopted spectral signatures of material surfaces in the short wave infrared range to distinguish human skin from other materials, regardless of the hide characteristics.
Bhattacharjee\etal\cite{bhattacharjee2017you} used off-the-shelf imaging devices to detect print and replay intrusions as well as mask-based presentation attacks near-real-time applications.
Many literature methods end up being restricted to specific datasets domains, especially when cameras have comparable capture quality.
However, they usually fail to achieve good results on cross-dataset evaluations~\cite{pinto2015using,boulkenafet2016face,liu2018learningSIW}.

  \section{The SWAX Dataset}
\label{sec:dataset}

{Realistic wax-made sculptures are susceptible to the same limitations found in silicon masks.
However, the former tends to be more faithful to actual human traits and, consequently, present greater chances of deceiving biometric systems than masks.
They can also be as portable as masks since presentation attacks may require either face portraits or busts instead of full-body figures.
From this perspective, we propose the \datasetName~(\datasetAbbr)\footnote{\scriptsize The \datasetAbbr~dataset will be released upon paper acceptance.}.}

The \datasetAbbr~database is compiled from unrestrained online resources and consists of characters, celebrities and public figure images and videos to whom wax dummies have been sculpted into.
{The database contains 33 female and 22 male individuals.
It consists of 1,812 images and 110 videos of 55 people/figures, arranged in training, validation and test sets. 
More precisely, each subject holds at least 20 authentic still images and a minimum of 10 counterfeit images.}
All motion and still pictures are manually captured under uncontrolled scenarios, formed by uncooperative individuals and distinct camera viewpoints.

Figure~\ref{fig:bona-wax-faces} points up the main aspects of the proposed database, depicting authentic and attack samples, respectively.
\textit{Bona fide} individual samples also incorporate human appearance alterations due to the natural aging process as pictures are collected without any sort of age restraint.
Some subjects samples also present accessories like spectacles or hats, and some male individuals may even have mustaches or beards.
Consequently, there is a significant variation in facial expressions, illumination, pose, scene, camera types and imaging conditions.

The proposed dataset comprises mutually exclusive sets, which indicates that no subject identity contained in the training set can be made available in the validation or the test sets simultaneously.
The same applies to validation and test sets, suggesting that alike subjects should not come out in distinct subsets. 
For development purposes, the proposed dataset encompasses independent randomly generated splits for 10-fold cross validation in an attempt to escape unfairly algorithm biases and expose overfitting occurrences.
In addition, the data presented in the \datasetAbbr~dataset should be used as is, according to the provided training, evaluation and test sets.

\subsection{Evaluation Protocols}

The following paragraphs specify the evaluation protocols guideline and set the proper manipulation of training, validation and test collections.
These sets are characterized by the corresponding subject/figure identity in which each face picture\footnote{\scriptsize The word ``picture'' refers to still images and videos.} is presented as either \textit{bona fide} (authentic) or \textit{counterfeit} (attack) sample. 
The four protocols\footnote{\scriptsize Protocol 01 follows the unsupervised paradigm whereas the remaining ones adhere to the supervised learning task.} are detailed in the following subsections.

\subsubsection{Protocol 01: Unsupervised, with additional data} 

Real-world biometric applications are inclined to anticipate all sorts of illegal intrusions and undergo attacks of distinct nature, which are unknown to the training stage.
However, unpredictable attack characteristics require a notable generalization potential that is not usually represented in supervised and multi-class classification techniques as they tend to become too specific to some particular attacks.

It is pertinent to handle spoofing detection problems with one-class-based techniques when labels are not available.
Alternate unsupervised strategies comprehend autoencoders~\cite{jourabloo2018face}, clustering~\cite{nikisins2018effectiveness} and domain adaptation~\cite{li2018unsupervised} algorithms.
One-class classification can be defined as a special case of unsupervised learning where only the class comprising authentic face pictures is well characterized by training data instances. 
It implies that counterfeit pictures are not known at training time but may emerge at test time.
In essence, having all picture samples sharing identical labels in the target variable is analogous to being unlabeled seeing that there is no discriminative information about their class names~\cite{liu2014unsupervised}. 

For the sake of following the unsupervised protocol appropriately, the following conditions shall apply:
\begin{itemize}[noitemsep]
    \item Procedures claimed to be unsupervised cannot make use of presentation attack samples at the training stage, being restricted to authentic samples only.
    \item There should be no parameters carrying \textit{bona fide} or \textit{counterfeit} labels, not to mention any other relevant information such as file names or singular identifiers.
    \item Approaches can neither benefit from ``beforehand information'' concerning the number of samples included in each class nor use the label distribution inherent to training and test sets.
    \item Supplementary samples, outside of \datasetAbbr~database, are allowed in cases where they depict \textit{bona fide} individuals only and should not be composed of hand-labeled data or information indicating whether pictures are authentic or comprise presentation attacks.
\end{itemize}

\subsubsection{Protocol 02: Restricted, without additional data}

The second protocol is in consonance with the supervised learning task, in which there is access to target variables for a set of training data.
Each face picture is provided with one out of two categories: \textit{bona fide} or \textit{counterfeit} class.
These two labels represent ground-truth information that favor the model in encountering the best-fitted mapping function to guarantee good predictions about ``future'' picture presentations.
That is, authentic and attack samples not listed in the training set to provide more accurate guidance on how well approaches generalize to unseen data.

In contrast to the unsupervised paradigm, this protocol admits information indicating whether a picture is authentic or consists of an attack.
Still, it dismisses any type of annotation or data from outside \datasetAbbr~database, including supplementary picture samples, external tools like facial landmark detectors and alignment methods learned on separate pictures, or feature extractors trained on other data sources. 
Such external algorithms must be unsupervised, and the data they operate on must be entirely within \datasetAbbr~database's training sets.
In other words, authentic/attack training labels provided in protocol 02 are authorized to be used along with external algorithms, provided that they satisfy the requirements below:
\begin{itemize}
    \item Algorithms under this protocol are not allowed to use supplementary data, either to identify presentation attacks or perform any kind of picture preprocessing.
    \item Validation and test sets are exclusive to their own purposes and cannot be employed to learn auxiliary methods.
    \item Researchers cannot rely on supplementary labeled data, such as manual face segmentation or facial landmark annotation.
\end{itemize}

\subsubsection{Protocol 03: Unrestricted, with no-label additional data}

Differently from the first two protocols, which restrain investigators and developers in such a way they must employ either no-label or \datasetAbbr~training data alone, protocol 03 acknowledges the exploitation of additional data sources in the interest of improving an algorithm's precision.
Actually, the main distinction between protocols 02 and 03 is that the latter supports the adoption of \datasetAbbr~independent data if and only if they do not consist of \textit{bona fide} or \textit{counterfeit} labels.

The external training data may encompass genuine or fraudulent face samples, but the annotation data must be limited to labeled or segmented face components, for instance, mouth contour and eye corners to name a few.
Auxiliary algorithms proposed for feature description, face detection and face alignment, even though having the possibility of being tuned on non-\datasetAbbr~face samples, are legitimate as we understand that outside data and methods used exclusively for no-spoofing reasons is significantly different from using supplementary data/methods to learn spoofing detection classifiers, deserving an exclusive category. 
Therefore, this protocol grants permission to utilizing feature extractors and alignment algorithms that have been developed for completely unrelated purposes.

The third protocol is distinguished from the others as it is subjected to the following conditions: 
\begin{itemize}
    \item Outside data should not consist of individuals included in \datasetAbbr~database.
    \item External pictures cannot hold corresponding information indicating whether they are genuine or fraudulent samples.
    \item Additional pictures can be labeled with keypoints or segments for the sake of designing preprocessing algorithms. 
    \item Non-\datasetAbbr~annotations cannot present information that may accredit the formation of authentic or attack face pictures.
\end{itemize}

\subsubsection{Protocol 04: Totally Unrestricted}
Supplementary labeled data play an important role in machine learning algorithms as it qualifies biometric algorithms to identify a broader variety of spoofing patterns.
On the contrary previously described protocols, protocol 04 endorses the use of additional picture samples regardless of the data annotation provided. 
The totally unrestricted pattern is the most permissive protocol as it admits outside datasets, external feature extractors and other methods that have been built on independent visual data as long as they adhere to the subsequent requirements:
\begin{itemize}
    \item Supplementary genuine and fraudulent pictures in which their corresponding identity is not available in the \datasetAbbr~database.
    \item External face samples may include annotated keypoints, attributes, segments as well as information carrying \textit{bona fide} or \textit{counterfeit} labels.
\end{itemize}

Protocol 04 admits cross-dataset experiments in the interest of assessing the generalization capability of algorithms and increasing their performance.
In consequence, researchers are allowed to avoid \datasetAbbr~samples, using external data only in the training stage, but compelled to evaluate the designed approaches on the provided testing splits. 
The idea of combining multiple datasets provides improved statistical power and enhanced classification ability on different domains, turning an algorithm more robust to diversified attacks.

\subsection{Evaluation Metrics}
Evaluation metrics are employed in order to point out a model's performance.
For the \datasetAbbr~database, we select two different metrics: the Receiver Operating Characteristic and the standardized ISO/IEC 30107-3 assessment mechanisms for biometric systems.

{The IEC 30107-3 designates a particular set of metrics for supervised and unsupervised paradigms~\cite{2017isobiometrics}.}
They are denominated Attack Presentation Classification Error Rate, $\text{\textsc{apcer}} = \frac{1}{V_{PA}} \sum_{i=1}^{V_{PA}}(1 - Res_{i})$; and Bona Fide Presentation Classification Error Rate, $\text{\textsc{bpcer}} = \frac{1}{V_{BF}} \sum_{i=1}^{V_{BF}}(Res_{i})$.
$V_{PA}$ indicates the number of spoofing attacks whereas $V_{BF}$ expresses the total number of authentic presentations.
$Res_{i}$ receives the value $1$ when the $i$-th probe video presentation is categorized as an attack and $0$ if labeled as \textit{bona fide} presentation.
\textsc{apcer} and \textsc{bpcer} resemble False Acceptance and False Rejection Rates, traditionally employed in the literature when assessing binary classification methods.

The Average Classification Error Rate (\textsc{acer}) summarizes the overall system performance as it comprehends the mean of \textsc{apcer} and \textsc{bpcer} at the decision threshold determined on the validation set~\cite{boulkenafet2017competition}.
Although not used in this work, researchers can also consider the Receiver Operating Characteristic (\textsc{roc}) and its associated Area Under Curve (\textsc{auc})~\cite{jain2000biometric} for the unsupervised task.

  \section{Baseline Experiments}
\label{sec:experiments}
This section analyzes the performance of some methods on the \datasetAbbr~benchmark and also details the employed feature descriptors, additional dataset and straightforward spoofing baselines.

\subsection{External Datasets}
Protocol 01 and 04 authorize researchers to incorporate external data. 
We select \textsc{lfw}~\cite{huang2008labeled} database, initially designed for studying the problem of unconstrained face recognition, to compose the unsupervised analysis.
\textsc{lfw} is compatible with \datasetAbbr's first protocol as it only comprises face images of distinct public figures, having no incidence of waxwork models. 
\textsc{mfsd}~\cite{wen2015face} satisfies the totally unrestricted protocol requirements as it provides authentic/attack label information for the training stage.

\subsection{Feature Descriptors}
We select \textsc{glcm}~\cite{haralick1973textural} and \textsc{lbp}~\cite{ledoux2016color} for the extraction of visual features in lower dimensional spaces.
\textsc{glcm} features are computed with directions $\theta \in \{0, 45, 90, 135\}$ degrees, distance $d \in \{1,2\}$, 16 bins and six texture properties: contrast, dissimilarity, homogeneity, energy, correlation, and angular second moment.
They are obtained from each image's corresponding Fourier transform in the frequency domain~\cite{pinto2015using}.
\textsc{lbp} information comprises 256 bins, a radius equal to 1, and eight points arranged in a $3\times3$ matrix.
They are taken from each image's color band after being converted from \textsc{rgb} into \textsc{hsv} and \textsc{yc\textsubscript{r}c\textsubscript{b}}~\cite{boulkenafet2015face}.
\textsc{all} refers to experiments in which \textsc{glcm} and \textsc{lbp} descriptors are extracted and concatenated. 

\subsection{Baselines}
As showed in Table~\ref{tab:baselines}, some approaches have been picked out from literature and employed as baselines. 
\textsc{Wsvm}~\cite{scheirer2014prob} is a classification algorithm that provides solutions for non-linear classification in open-set scenarios.
For the unsupervised paradigm, we combine one-class \textsc{Wsvm} with different feature descriptors. 

\textsc{Rhythm}~\cite{pinto2015using} searches for frequency domain artifacts as it is trained and tested on \datasetAbbr's second and third protocols in favor of distinguishing counterfeit from valid images.
Since Protocol 03 admits external data and methods as long as they are not designed for spoofing purposes, additional algorithms can be engaged in preprocessing tasks.
\textsc{Align}~\cite{kowalski2017deep} is a \textsc{dnn}-based face alignment method that iteratively improves the locations of the facial landmarks estimated in previous stages. 
Such alignment algorithms tend to improve biometric systems performance due to carefully positioning subject faces into a canonical pose.
For Protocol 04, \textsc{De-Spoofing}~\cite{jourabloo2018face} performs a cross-dataset evaluation as it is trained on \textsc{oulu-npu} dataset~\cite{boulkenafet2017oulu} and then aims to estimate deep spoof noises from \datasetAbbr~probe face samples.

We propose an embedding of \textsc{pls}~\cite{rosipal2005overview} and \textsc{svm}~\cite{steinwart2008support} classifiers, entitled \textsc{Epls} and \textsc{Esvm}, respectively.
The embedding is learned on random subsets of the training set to create an array of classifiers, guaranteeing a balanced division of \textit{bona fide} (positive class) and \textit{counterfeit} (negative class) samples within each classification model.
At test time the method projects a probe image onto all trained classifiers and computes the ratio of the number of positive responses attained to the total number of classification models.
If most classifiers return positive responses, it implies that the face image is likely to be an authentic sample, or a spoofing attack, otherwise.

\begin{table}[!t]
  \centering
  \scriptsize
  \caption{\textsc{apcer} and \textsc{bpcer} results (\%) on \datasetAbbr's protocols. Supplementary datasets are depicted  between parenthesis.}
  \begin{tabular}{|c|r|c|c|c|} \hline                                            
    \tiny{\textbf{Prot.}} & \tiny{\textbf{Method}}  & \textsc{apcer} & \textsc{bpcer} & \textsc{acer}   \\ \hline \hline
    \multirow{4}{*}{1} & \textsc{glcm+Wsvm}         & $85.8\pm3.65 $ & $20.2\pm3.93 $ & $53.0\pm3.79 $  \\ 
                       & \textsc{lbp+Wsvm}          & $84.2\pm4.92 $ & $11.4\pm2.15 $ & $47.8\pm3.53 $  \\ 
                       & \textsc{all+Wsvm}          & $89.9\pm3.10 $ & $11.0\pm4.26 $ & $50.5\pm3.68 $  \\ 
                       & \textsc{all+Wsvm(lfw)}     & $83.6\pm4.38 $ & $16.2\pm5.32 $ & $49.9\pm4.85 $  \\ \hline \hline
    \multirow{3}{*}{2} & \textsc{Rhythm}            & $57.8\pm12.1 $ & $41.0\pm8.84 $ & $49.5\pm10.5 $  \\ 
                       & \textsc{all+Epls}          & $34.6\pm1.75 $ & $20.6\pm2.13 $ & $27.6\pm1.94 $  \\
                       & \textsc{all+Esvm}          & $46.0\pm2.12 $ & $19.9\pm2.10 $ & $33.0\pm2.11 $  \\ \hline \hline
    \multirow{3}{*}{3} & \textsc{Align+Rhythm}      & $51.1\pm5.24 $ & $43.2\pm4.86 $ & $47.1\pm5.05 $  \\ 
                       & \textsc{Align+all+Epls}    & $32.4\pm1.68 $ & $19.8\pm1.25 $ & $26.1\pm1.46 $  \\
                       & \textsc{Align+all+Esvm}    & $44.9\pm3.29 $ & $22.0\pm3.40 $ & $33.5\pm3.34 $  \\ \hline \hline
    \multirow{3}{*}{4} & \textsc{De-Spoofing}       & $13.6\pm6.20 $ & $67.6\pm11.9 $ & $40.6\pm9.07 $  \\
                       & \textsc{all+Epls(mfsd)}    & $32.1\pm1.21 $ & $20.1\pm1.41 $ & $26.1\pm1.31 $  \\
                       & \textsc{all+Esvm(mfsd)}    & $47.9\pm1.75 $ & $17.6\pm1.11 $ & $32.7\pm1.43 $  \\ \hline
  \end{tabular}
  \label{tab:baselines}
\end{table}

Evaluated algorithms provide an insight into the performance of countermeasure methods when confronted with wax figure images. 
Experiments in Table~\ref{tab:baselines} show that advanced spoofing methods are still vulnerable to high-quality violation attempts, demanding researchers to deliver more accurate biometric systems. 
The attained results are not satisfactory and reveal that face anti-spoofing has a large room for improvement despite the significant progress in recent years.

  \section{Conclusions}
\label{sec:conclusion}

This work comprises a public benchmark of real human and wax figure images and videos, entitled \datasetName, developed for face anti-spoofing purposes.
The database consists of different protocols, cross validation splits and evaluation metrics to operate fair comparisons among different algorithms. 
Results show that wax figures can be employed in pursuance of bypassing biometric systems and obtain illegitimate access.
Therefore, there is a lot to improve when it comes to wax-based face spoofing attacks.
We hope that with the advent of this dataset researchers will consider new sorts of spoofing attacks as a motivation to delivering robust methods to the challenging area of spoofing detection.


  {
    \balance
    \small
    \bibliographystyle{IEEEbib}
    \bibliography{references}
  }

\end{document}